\documentclass[runningheads]{llncs}
\usepackage{graphicx}
\usepackage{multirow}
\usepackage{booktabs}
\usepackage{makecell}
\usepackage{threeparttable}
\usepackage{makecell}
\usepackage{amsmath}
\usepackage{amsfonts}
\usepackage{graphicx}

\begin{document}

\title{Double-Uncertainty Guided Spatial and Temporal Consistency Regularization Weighting for Learning-based Abdominal Registration}

\titlerunning{Double-Uncertainty Guided Regularization Weighting}
%
\author{Zhe Xu\inst{1} \and Jie Luo\inst{3} \and
Donghuan Lu\inst{2} \and
Jiangpeng Yan\inst{4} \and  Sarah Frisken\inst{3} \and Jayender Jagadeesan\inst{3} \and William Wells III\inst{3} \and   Xiu Li\inst{5} \and Yefeng Zheng\inst{2} \and Raymond Tong\inst{1}}

\authorrunning{Z. Xu et al.}
%
\institute{Department of Biomedical Engineering, The Chinese University of Hong Kong, Hong Kong, China
\and Tencent Jarvis Lab, Shenzhen, China\\
\and Brigham and Women’s Hospital, Harvard Medical School, Boston, USA \\
\and Department of Automation, Tsinghua University, Beijing, China \\
\and Shenzhen International Graduate School, Tsinghua University, Shenzhen, China \\
}

\maketitle              
\begin{abstract} 
In order to tackle the difficulty associated with the ill-posed nature of the image registration problem, regularization is often used to constrain the solution space. For most learning-based registration approaches, the regularization usually has a fixed weight and only constrains the spatial transformation. Such convention has two limitations: (i) Besides the laborious grid search for the optimal fixed weight, the regularization strength of a specific image pair should be associated with the content of the images, thus the ``one value fits all'' training scheme is not ideal; (ii) Only spatially regularizing the transformation may neglect some informative clues related to the ill-posedness. In this study, we propose a mean-teacher based registration framework, which incorporates an additional temporal consistency regularization term by encouraging the teacher model's prediction to be consistent with that of the student model. More importantly, instead of searching for a fixed weight, the teacher enables automatically adjusting the weights of the spatial regularization and the temporal consistency regularization by taking advantage of the transformation uncertainty and appearance uncertainty. Extensive experiments on the challenging abdominal CT-MRI registration show that our training strategy can promisingly advance the original learning-based method in terms of efficient hyperparameter tuning and a better tradeoff between accuracy and smoothness.

\end{abstract}
\section{Introduction}
Recently, learning-based multimodal abdominal registration has greatly advanced percutaneous nephrolithotomy due to their substantial improvement in computational efficiency and accuracy \cite{VM2018,xu2020adversarial,hering2021learn2reg,de2019deep}. In the training stage, given a set of paired image data, the neural network optimizes the cost function
\begin{equation}
\label{eq1}
\mathcal{L} =\mathcal{L}_{sim}\left(I_{f}, I_{m} \circ \phi\right)+\lambda \mathcal{L}_{reg}(\phi),
\end{equation}
to learn a mapping function that can rapidly estimate the deformation field $\phi$ for a new pair of images. In the cost function, the first term $\mathcal{L}_{sim}$ quantifies the appearance dissimilarity between the fixed image $I_{f}$ and the warped moving image $I_{m} \circ \phi$. Since image registration is an ill-posed problem, the second regularization term $\mathcal{L}_{reg}$ is used to constrain its solution space. The tradeoff between registration accuracy and transformation smoothness is controlled by a weighting coefficient $\lambda$.

\begin{figure}[t]
\includegraphics[width=\textwidth]{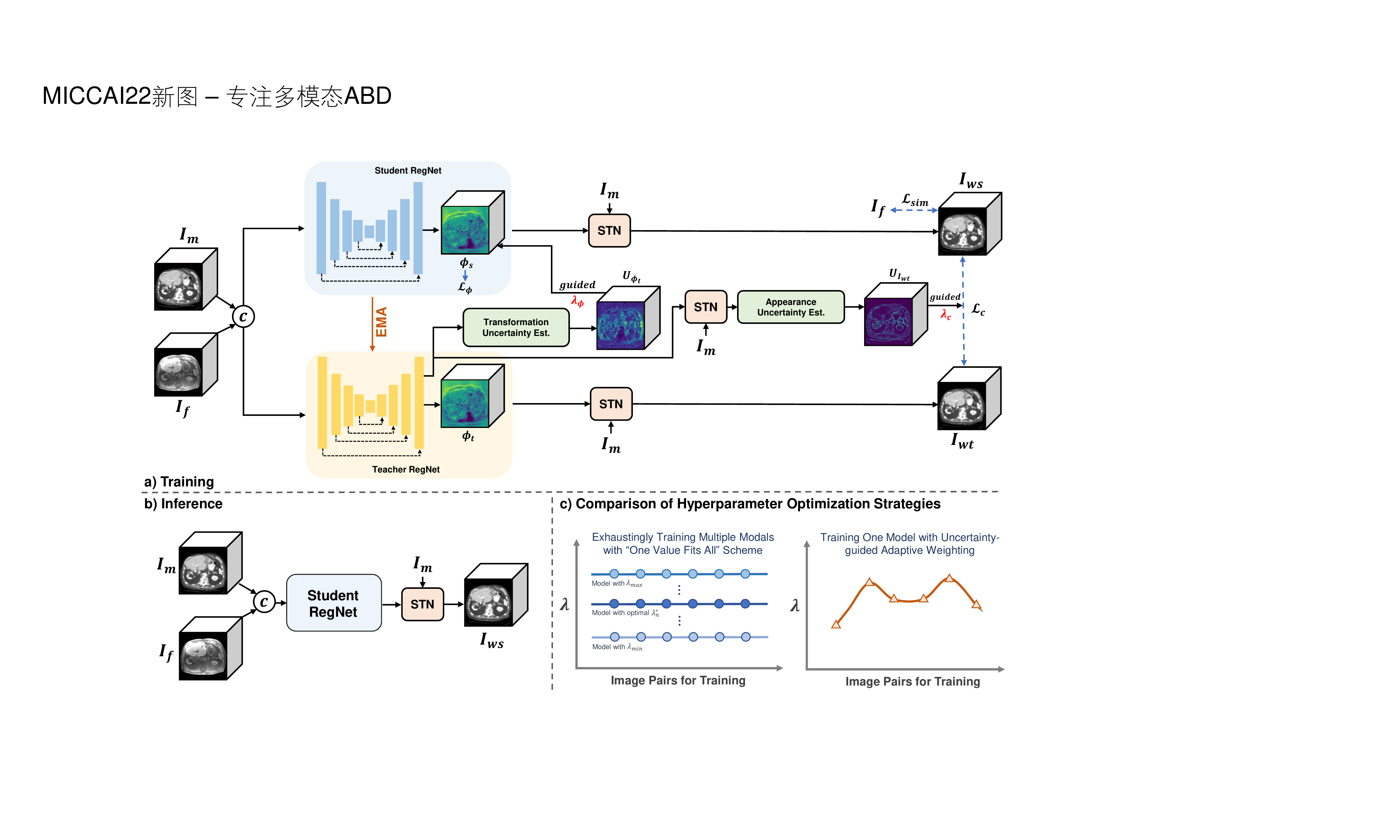}
\caption{Illustration of the proposed framework with (a) training process, (b) inference process and (c) comparison of hyperparameter optimization strategies.} \label{fig1}
\end{figure}

In classical iterative registration methods, the weight $\lambda$ is manually tuned for each image pair. Yet, in most learning-based approaches \cite{VM2018}, the weight $\lambda$ is commonly set to a fixed value for all image pairs throughout the training stage, assuming that they require the same regularization strength. Besides the notoriously time-consuming grid search for the so-called ``optimal" fixed weight, this ``one value fits all'' scheme (Fig. \ref{fig1}(c)) may be suboptimal because the regularization strength of a specific image pair should be associated with their content and misalignment degree, especially for the challenging abdominal registration with various large deformation patterns. In these regards, HyperMorph \cite{hoopes2021hypermorph} estimated the effect of hyperparameter values on deformations with the additionally trained hypernetworks \cite{ha2016hypernetworks}. Being more parameter-efficient, recently, Mok et al. \cite{mok2021conditional} introduced conditional instance normalization \cite{dumoulin2016learned} into the backbone network and used an extra distributed mapping network to implicitly control the regularization by normalizing and shifting feature statistics with their affine parameters. As such, it enables optimizing the model with adaptive regularization weights during training and reducing the human efforts in hyperparameter tuning. We also focus on this underexploited topic, yet, propose an explicit and substantially different alternative without changing any components of the backbone.

On the other hand, besides the traditional regularization terms, e.g., smoothness and bending energy \cite{Bookstein}, task-specific regularizations have been further proposed, e.g., population-level statistics \cite{bhalodia2019cooperative} and biomechanical models \cite{hu2018adversarial,qin2020biomechanics}. Inherently, all these methods still belong to the category of \textit{spatial regularization}. Experimentally, however, we notice that the estimated solutions in unsupervised registration vary greatly at different training steps due to the ill-posedness. Only spatially regularizing the transformation at each training step may neglect some informative clues related to the ill-posedness. It might be advantageous to further exploit such temporal information across different training steps. 

In this paper, we present a novel double-uncertainty guided spatial and temporal consistency regularization weighting strategy, in conjunction with the Mean-Teacher (MT) \cite{tarvainen2017mean} based registration framework. Specifically, inspired by the consistency regularization \cite{tarvainen2017mean,wu2021semi} in semi-supervised learning, this framework further incorporates an additional \textit{temporal consistency regularization} term by encouraging the consistency between the predictions of the student model and the temporal ensembling predictions of the teacher model. More importantly, instead of laboriously grid searching for the optimal fixed regularization weight, the self-ensembling teacher model takes advantage of the transformation uncertainty and the appearance uncertainty \cite{Jax} derived by Monte Carlo dropout \cite{MCD} to heuristically adjust the regularization weights for each image pair during training (Fig. \ref{fig1} (c)). Extensive experiments on a challenging intra-patient abdominal CT-MRI dataset show that our training strategy can promisingly advance the original learning-based method in terms of efficient hyperparameter tuning and a better tradeoff between accuracy and smoothness.


\section{Methods}
The proposed framework, as depicted in Fig. \ref{fig1} (a), is designed on top of the Mean-Teacher (MT) architecture \cite{tarvainen2017mean} which is constructed by a student registration network (Student RegNet) and a weight-averaged teacher registration network (Teacher RegNet). This architecture was originally proposed for semi-supervised learning, showing superior performance in further exploiting unlabeled data. We appreciate the MT-like design for registration because the temporal ensembling strategy can (i) help us efficiently exploit the temporal information across different training steps in such a ill-posed problem (Sec. \ref{Temporal}), and (ii) enable smoother uncertainty estimation \cite{yu2019uncertainty} to heuristically adjust the regularization weights during training (Sec. \ref{weighting}).

\subsection{Mean-Teacher based Temporal Consistency Regularization}
\label{Temporal}
Specifically, the student model is a typical registration network updated by back-propagation, while the teacher model uses the same network architecture as the student model but its weights are updated from that of the student model via Exponential Moving Average (EMA) strategy, allowing to exploit the temporal information across the adjacent training steps. Formally, denoting the weights of the teacher model and the student model at training step $k$ as $\theta_{k}^{\prime}$ and $\theta_{k}$, respectively, we update $\theta_{k}^{\prime}$ as: 
\begin{equation}
\theta_{k}^{\prime}=\alpha \theta_{k-1}^{\prime}+(1-\alpha) \theta_{k},
\end{equation}
where $\alpha$ is the EMA decay and empirically set to $0.99$ \cite{tarvainen2017mean}. To exemplify our paradigm, we adopt the same U-Net architecture used in VoxelMorph (VM) \cite{VM2018} as the backbone network. Concisely, the moving image $I_{m}$ and the fixed image $ I_{f}$ are concatenated as a single 2-channel 3D image input, and downsampled by four $3 \times 3 \times 3$ convolutions with stride of 2 as the encoder. Then, corresponding 32-filter convolutions and four upsampling layers are applied to form a decoder, followed by four convolutions to refine the 3-channel deformation field. Skip connections between encoder and decoder are also applied. Given the predicted $\phi_{s}$ and $\phi_{t}$ from the student and the teacher, respectively, the Spatial Transformation Network (STN) \cite{STN} is utilized to warp the moving image $I_{m}$ into $I_{ws}$ and $I_{wt}$, respectively. Following the common practice, the dissimilarity between $I_{ws}$ and $I_{f}$ can be measured as the similarity loss $\mathcal{L}_{sim}(I_{ws},I_{f})$ with a spatial regularization term $\mathcal{L}_{\phi}$ for $\phi_{s}$. To exploit the informative temporal information related to the ill-posedness, we encourage the temporal ensemble prediction of the teacher model to be consistent with that of the student model by adding an appearance consistency constraint $\mathcal{L}_{c}(I_{ws},I_{wt})$ to the training loss. In contrast to $\mathcal{L}_{\phi}$ which spatially constrains the deformation field at the current training step, $\mathcal{L}_{c}$ penalizes the difference between predictions across adjacent training steps. Thus, we call $\mathcal{L}_{c}$ \textit{temporal consistency regularization}.

As such, the cost function Eqn. \ref{eq1} can be reformulated as the following Eqn. \ref{eq3}, which is a combination of similarity loss $\mathcal{L}_{sim}$, spatial regularization loss $\mathcal{L}_{\phi}$ and temporal consistency regularization loss $\mathcal{L}_{c}$:
\begin{equation}
\label{eq3}
\mathcal{L} = \mathcal{L}_{sim}+\lambda_{\phi} \mathcal{L}_{\phi} + \lambda_{c} \mathcal{L}_{c},
\end{equation}
where $\lambda_{\phi}$ and $\lambda_{c}$ are the uncertainty guided adaptive tradeoff weights, as elaborated in the following Sec. \ref{weighting}. In order to handle multimodal abdominal registration, we use the Modality Independent Neighborhood Descriptor (MIND) \cite{mind} based dissimilarity metric for $\mathcal{L}_{sim}$. $\mathcal{L}_{c}$ is measured by mean squared error (MSE). Following the benchmark method \cite{VM2018}, the choice of $\mathcal{L}_{\phi}$ is the generic L2-norm of the deformation field gradients for smooth transformation. 

\subsection{Double-Uncertainty Guided Adaptive Weighting}
\label{weighting}
Distinct from the previous ``one value fits all'' strategy, we propose a double-uncertainty guided adaptive weighting scheme to locate the uncertain samples and then adaptively adjust $\lambda_{\phi}$ and $\lambda_{c}$ for each training step.\\

\noindent\textbf{Double-Uncertainty Estimation.} 
Besides predicting the deformation $\phi_{t}$, the teacher model can also serve as our uncertainty estimation branch since the model weight-averaged strategy can improve the stability of the predictions \cite{tarvainen2017mean} that enables smoother model uncertainty estimation \cite{yu2019uncertainty}. Particularly, we adopt the well-known Monte Carlo dropout \cite{MCD} for Bayesian approximation due to its superior robustness \cite{qu2021dat}. We repetitively perform $N$ stochastic forward passes on the teacher model with random dropout. After this step, we can obtain a set of voxel-wise predicted deformation fields $\left\{\phi_{t_i}\right\}_{i=1}^{N}$ with a set of warped images $\left\{I_{wt_i}\right\}_{i=1}^{N}$. We propose to use \textit{the absolute value of the ratio of standard deviation to the mean}, which can characterize the normalized volatility of the predictions, to represent the uncertainty \cite{smith2018understanding}. More specifically, the proposed registration uncertainties can be categorized into the transformation uncertainty and appearance uncertainty \cite{Jax}. Formulating $\mu_{\phi_{t}}^{c}=\frac{1}{N} \sum_{i=1}^{N} \phi_{t_i}^{c}$ and $\mu_{I_{wt}}=\frac{1}{N} \sum_{i=1}^{N} I_{wt_i}$ as the mean of the deformation fields and the warped images, respectively, where $c$ represents the $c^{th}$ channel of the deformation field (i.e., $x$, $y$, $z$ displacements) and $i$ denotes the $i^{th}$ forward pass, the transformation uncertainty map $U_{\phi_{t}} \in \mathbb{R}^{H \times W \times D \times 3}$ and the appearance uncertainty map $U_{I_{wt}} \in \mathbb{R}^{H \times W \times D}$ can be calculated as:
\begin{equation}
\left\{\begin{array}{l}
\sigma_{\phi_{t}}^{c}=\sqrt{\frac{1}{N-1} \sum_{i=1}^{N}\left(\phi_{t_i}^{c}-\mu_{\phi_{t}}^{c}\right)^{2}} \quad \text { and } \quad U_{\phi_{t}}^{c}=\left |\frac{\sigma_{\phi_{t}}^{c}}{\mu_{\phi_{t}}^{c}}\right|, \\
\sigma_{I_{wt}}=\sqrt{\frac{1}{N-1} \sum_{i=1}^{N}\left(I_{wt_i}-\mu_{I_{wt}}\right)^{2}} \quad \text { and } \quad U_{I_{wt}}=\left|\frac{\sigma_{I_{wt}}}{\mu_{I_{wt}}}\right|.
\end{array}\right.
\end{equation}

With the guidance of $U_{\phi_{t}}$ and $U_{I_{wt}}$, we then propose to heuristically assign the weights $\lambda_{\phi}$ and $\lambda_{c}$ of the spatial regularization $\mathcal{L}_{\phi}$ and the temporal consistency regularization $\mathcal{L}_{c}$ for each image pair during training.\\

\noindent\textbf{Adaptive Weighting.} 
Firstly, for the typical spatial regularization $\mathcal{L}_{\phi}$, considering that unreliable predictions often correlate with biologically-implausible deformations \cite{xu2020unimodal}, we assume that stronger spatial regularization can be given when the network tends to produce more uncertain predictions. As for the temporal consistency regularization $\mathcal{L}_{c}$, we notice that more uncertain predictions can be characterized as that this image pair is harder-to-align. Particularly, the most recent work \cite{wu2021semi} in semi-supervised segmentation combats with the assumption in \cite{yu2019uncertainty} and experimentally reveals an interesting finding that emphasizing the unsupervised teacher-student consistency on those unreliable (often challenging) regions can provide more informative and productive clues for network training. Herein, we follow this intuition, i.e., more uncertain (difficult) samples should receive more attention (higher $\lambda_{c}$) for the consistency regularization $\mathcal{L}_{c}$. Formally, for each training step $s$, we update $\lambda_{\phi}$ and $\lambda_{c}$ as follows:
\begin{equation}
\lambda_{\phi}(s)=k_{1} \cdot \frac{\sum_{v} \mathbb{I}\left(U_{\phi_{t}}(s)>\tau_{1}\right)}{V_{U_{\phi_{t}}}} \quad \text { and } \quad \lambda_{c}(s)=k_{2} \cdot \frac{\sum_{v}\mathbb{I}\left(U_{I_{wt}}(s)>\tau_{2}\right)}{V_{U_{I_{wt}}}},
\end{equation}
where $\mathbb{I}(\cdot)$ is the indicator function; $v$ denotes the $v$-th voxel; $V_{U_{\phi_{t}}}$ and $V_{I_{wt}}$ represent the volume sizes of $U_{\phi_{t}}$ and $U_{I_{wt}}$, respectively; $k_{1}$ and $k_{2}$ are the empirical scalar values; and $\tau_{1}$ and $\tau_{2}$ are the thresholds to select the most uncertain predictions. Noteworthily, the proposed strategy can work with any learning-based image registration architectures without increasing the number of trainable parameters. Besides, as shown in Fig. \ref{fig1} (b), only the student model is utilized at the inference stage, which can ensure the computational efficiency. 

\section{Experiments and Results}
\subsubsection{Datasets.} 
We focus on the challenging application of abdominal CT-MRI multimodal registration for improving the accuracy of percutaneous nephrolithotomy. Under institutional review board approval, a 50-pair intra-patient abdominal CT-MRI dataset was collected from our partnership hospital with radiologist-examined segmentation masks for the region-of-interests (ROIs), including liver, kidney and spleen. We randomly divided the dataset into three groups for training (35 cases), validation (5 cases) and testing (10 cases), respectively. After sequential preprocessing steps including resampling, affine pre-alignment, intensity normalization and cropping, the images were processed into sub-volumes of $176 \times$ $176 \times 128$ voxels at the 1 $mm$ isotropic resolution.\\

\noindent\textbf{Implementation and Evaluation Criteria.} 
The proposed framework is implemented on PyTorch and trained on an NVIDIA Titan X (Pascal) GPU. We employ the Adam optimizer with a learning rate of 0.0001 with a decay factor of 0.9. The batch size is set to 1 so that each step contains an image pair. We set $N=6$ for the uncertainty estimation. We empirically set $k_{1}$ to $5$ since the deformation fields with maximum $\lambda_{\phi}=5$ are diffeomorphic in most cases. The scalar value $k_{2}$ is set to $1$. Thresholds $\tau_{1}$ and $\tau_{2}$ are set to $10\%$ and $1\%$, respectively. We adopt a series of evaluation metrics, including Average Surface Distance (ASD) and the average Dice score between the segmentation masks of warped images and fixed images. In addition, the average percentage of voxels with non-positive Jacobian determinant ($|J_{\phi}| \leq 0$) in the deformation fields and the standard deviation of the Jacobian determinant ($\sigma(|J_{\phi}|)$) are obtained to quantify the diffeomorphism and smoothness, respectively.\\

\begin{figure}[t]
\centering
\includegraphics[width=0.98\textwidth]{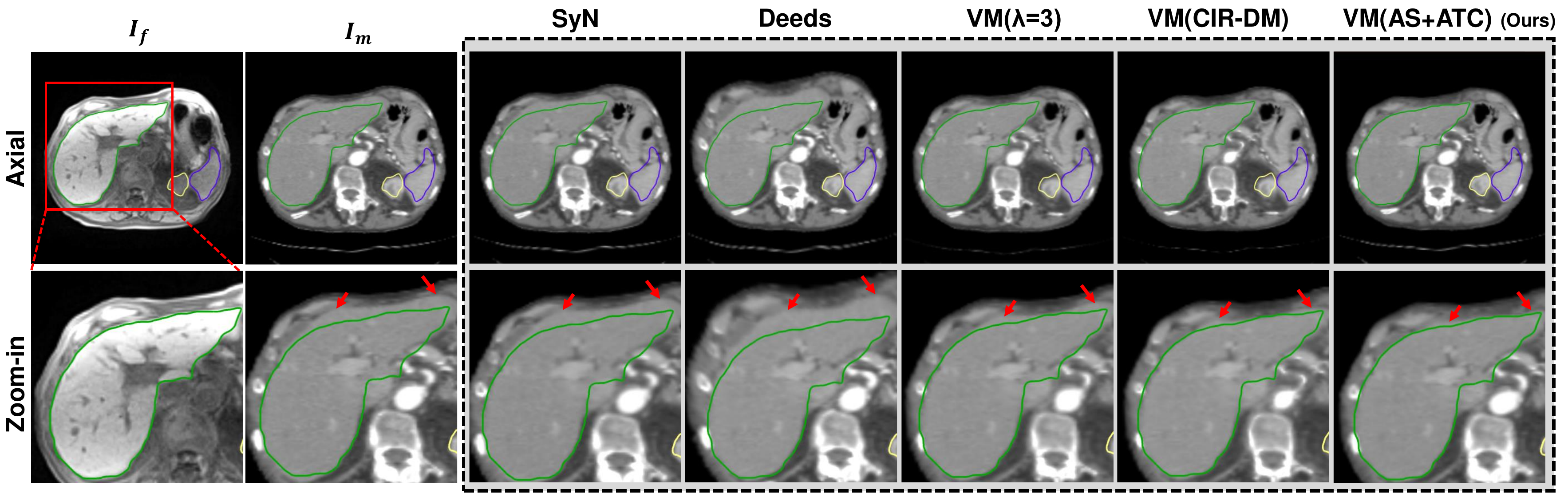}
\caption{Exemplar axial slice of an abdominal CT-MRI registration case. The segmentation contours of the liver (green), kidney (yellow) and spleen (blue) extracted from the fixed abdominal MRI $I_f$ are overlaid on all images. Better alignment drives structures closer to the fixed contours of $I_{f}$. The red arrows indicate the registration of interest around the organ boundary.} 
\label{fig_result}
\end{figure}

\noindent\textbf{Comparison Study.}
Table \ref{tab1} and Fig. \ref{fig_result} present the quantitative and qualitative comparisons, respectively. We compare with several baselines, including two traditional methods SyN \cite{Avants2008SymmetricDI} and Deeds \cite{heinrich2013mrf} with five levels of discrete optimization, as well as the benchmark learning-based method VoxelMorph (VM) \cite{VM2018} and its probabilistic diffeomorphic version DIF-VM \cite{VM-DIF}. As an alternative for adaptive weighting, we also include the recent conditional registration network \cite{mok2021conditional} with the same scalar value $5$ (denoted as VM (CIR-DM)). Fairly, we use the MIND-based dissimilarity metric in all learning-based methods. Although DIF-VM preserves better diffeomorphism properties, we find that its results are often suboptimal. Thus, we adopt VM as our backbone. For simplicity, we denote our adaptive spatial and temporal consistency regularization weighting strategy as VM (AS+ATC). Instead of training multiple models for finding the optimal fixed weight, only one model needs to be trained in both VM (CIR-DM) \cite{mok2021conditional} and our VM (AS+ATC). Experimentally, training each VM model from scratch requires an average of $9.2\mathrm{h}$ for this task. For the typical grid search scheme, five individual VM models are trained with varying fixed spatial regularization weights from $1$ to $5$, resulting in around $46 \mathrm{h}$ training time in total, wherein we observe that the overall best-performing VM model appears at $\lambda=3$. As a comparison, the implicit control method, i.e., VM (CIR-DM), achieves comparable results with $\sim$4.7x shorter total training time. Compared with VM (CIR-DM), our VM (AS+ATC) requires slightly longer training time due to the uncertainty estimation, yet, still resulting in $\sim$4.2x faster than the grid search scheme. Distinct from VM (CIR-DM), we do not need to change any components in the network. Encouragingly, we find that VM (AS+ATC) further improves the registration accuracy in terms of Dice and ASD along with better properties of diffeomorphism and smoothness, implying that our strategy helps produce more desirable (smoother) solutions in this ill-posed problem. Visually, the structure boundaries registered by SyN and Deeds still have considerable disagreements, while the learning-based methods achieve more appealing boundary alignment. Besides, all trained models can infer an alignment in a second with a GPU.\\

\begin{table}[t]\scriptsize
\caption{Quantitative results for abdominal CT-MRI registration (mean$\pm$std). Higher average Dice score (\%) and lower ASD (mm) are better.  $\dagger$ indicates the best model via grid search. Best results are shown in bold. Average percentage of foldings ($\left|J_{\phi}\right| \leq 0$) and the standard deviation of the Jacobian determinant ($\sigma(|J_{\phi}|)$) are also given.}\label{tab1}
\centering
\scalebox{0.84}{
\begin{tabular}{l|ccc|ccc|c|c}
\Xhline{1pt}
\multirow{2}{*}{Methods}                                                                                                                       
                         & \multicolumn{3}{c|}{Dice {[}\%{]} $\uparrow$}                          & \multicolumn{3}{c|}{ASD {[}voxel{]} $\downarrow$}                                 & \multirow{2}{*}{$\left|J_{\phi}\right| \leq 0$} & \multirow{2}{*}{$\sigma(|J_{\phi}|)$}  \\ \cline{2-7}
                         & Liver                 & Spleen                & Kidney & Liver                 & \multicolumn{1}{c}{Spleen} & \multicolumn{1}{c|}{Kidney} &                                                 &                                                                             \\ \Xhline{0.7pt}
Initial                  &         76.23$\pm$4.12              &    77.94$\pm$3.31                   &    80.18$\pm$3.06    &          4.98$\pm$0.83             &              2.02$\pm$0.51               &   1.95$\pm$0.35     &        -                                         &            -                                                       \\ 
SyN                      &        79.42$\pm$4.35               &   80.33$\pm$3.42                    &   82.68$\pm$3.03     &           4.83$\pm$0.82            &           1.62$\pm$0.61                  &    1.91$\pm$0.44    &                0.07\%                                 &            0.42                                                           \\ 
Deeds                    &        82.16$\pm$3.15               &    81.48$\pm$2.64                   &    83.82$\pm$3.02    &         3.97$\pm$0.55              &          1.44$\pm$0.62                   &    1.59$\pm$0.40    &    0.01\%                                             &            0.28                                                        \\ 
DIF-VM                    &        83.14$\pm$3.24               &          82.45$\pm$2.59             &    83.24$\pm$2.98    &          3.88$\pm$0.62             &        1.52$\pm$0.59                     &   1.63$\pm$0.41    &    \textless 0.001\%                                             &            0.12                                                       \\ \hline
VM ($\lambda = 1$)        &        85.21$\pm$3.06               &       84.04$\pm$2.52                &    83.12$\pm$2.81    &         3.17$\pm$0.59              &           1.34$\pm$0.56                  &     1.55$\pm$0.37   &            0.03\%                                     &            0.19                                                        \\ 
VM ($\lambda = 3$)$^\dagger$        &       85.36$\pm$3.12                &           84.24$\pm$2.61            &    83.40$\pm$3.11    &           3.19$\pm$0.53             &         1.31$\pm$0.52                    &    1.56$\pm$0.42    &   0.001\%                                              &     0.14                                                           \\ 
VM ($\lambda = 5$)        &         84.28$\pm$2.93              &          83.71$\pm$2.40             &  82.96$\pm$2.77      &        3.83$\pm$0.68               &        1.48$\pm$0.59                     &   1.65$\pm$0.44     &              \textless 0.0001\%                                   &          0.08                                                         \\\hline
VM (CIR-DM)         &       85.29$\pm$3.39                &           84.17$\pm$2.59            &    83.01$\pm$3.06    &           3.32$\pm$0.41             &         1.33$\pm$0.47                    &    1.49$\pm$0.46    &   0.002\%                                              &     0.17     \\
VM (AS+ATC) (ours)              & \textbf{87.01$\pm$3.22}   &    \textbf{84.96$\pm$2.55}   &        \textbf{84.47$\pm$2.86} &              \textbf{2.57$\pm$0.48}               &    1.24$\pm$0.49    &               \textbf{1.26$\pm$0.43}                                  &              \textless 0.0005\%  &   0.13                                                   \\\hline
\multicolumn{9}{c}{Ablation Study}\\\hline
VM (AS)                 &       86.02$\pm$3.02                &           84.75$\pm$2.59            &    83.44$\pm$2.96    &           3.21$\pm$0.49             &         1.33$\pm$0.50                    &    1.45$\pm$0.49    &   0.0007\%                                              &     0.12    \\ 
VM (S+TC)                 &       86.32$\pm$3.07                &           84.37$\pm$2.53            &    83.95$\pm$3.02    &           3.05$\pm$0.50             &         1.26$\pm$0.52                    &    1.47$\pm$0.47    &   0.001\%                                              &     0.15                                                     \\
VM (AS+TC)               & 86.49$\pm$3.13    & 84.87$\pm$2.68   &    84.03$\pm$2.89    & 2.88$\pm$0.53 &           \textbf{1.20$\pm$0.44}              &    1.32$\pm$0.40    &       0.0005\%                                         &          0.14                                                     \\
\Xhline{1pt}
\end{tabular}}
\end{table}

\noindent\textbf{Ablation Study.}
To better understand our training strategy, we perform an ablation study with three variants: (i) VM (AS): removing the adaptive temporal consistency term; (ii) VM (S+TC): using empirical fixed weights $\lambda_{\phi}=3$ and $\lambda_{c}=0.5$; (iii) VM (AS+TC): adaptively adjusting $\lambda_{\phi}$ while $\lambda_{c}$ is fixed as 0.5. The quantitative results are also presented in Table \ref{tab1}. Especially, similar to VM (CIR-DM), VM (AS) only adaptively adjusts $\lambda_{\phi}$. We find that VM (AS) achieves even better performance compared with VM ($\lambda=3$) and VM (CIR-DM), highlighting that our uncertainty-guided weighting scheme enables more precise control of the spatial regularization strength during training. It can be also observed that both VM (S+TC) and VM (AS+TC) achieve better performance than VM (AS), demonstrating that additionally exploiting the temporal information can be rewarding. When we integrate these components into our synergistic training scheme, their better efficacy can be brought into play.\\

\begin{figure}[t]
\centering
\includegraphics[width=0.95\textwidth]{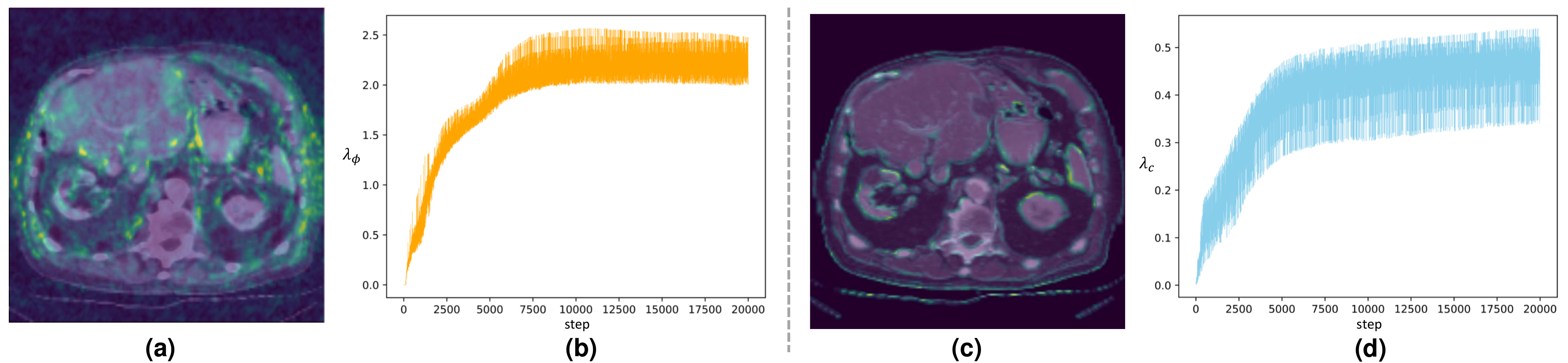}
\caption{(a) and (c) are the examples of $U_{\phi_{t}}$ and $U_{I_{wt}}$  (overlay on the moving image), respectively, where brighter areas denote more uncertain regions. (b) and (d) show the adaptive weighting process across the training steps for $\lambda_{\phi}$ and $\lambda_{c}$, respectively.} \label{fig_weighting}
\end{figure}

\noindent\textbf{Visualized Uncertainty Map and Weighting Process.}
Examples of two uncertainty maps $U_{\phi_{t}}$ and $U_{I_{wt}}$ are visualized in Fig. \ref{fig_weighting} (a) and (c), respectively. We observe that high uncertainty often occurs in hard-to-align ambiguous areas. Note that at the early stage, the weights are relatively small since limited transformation has been captured. As the training goes, the weights of $\mathcal{L}_{\phi}$ and $\mathcal{L}_{c}$ are adaptively modulated after each step (Fig. \ref{fig_weighting} (b) and (d)) assisted by the two uncertainties. Such scheme helps the model pursue a better tradeoff between accurate alignment and desirable diffeomorphism properties.\\

\section{Conclusion}
In this paper, we proposed a double-uncertainty guided spatial and temporal consistency regularization weighting strategy, assisted by a mean-teacher based registration framework. Besides temporal consistency regularization for further exploiting the temporal clues related to the ill-posedness, more importantly, the self-ensembling teacher model takes advantage of two estimated uncertainties to heuristically adjust the regularization weights for each image pair during training. Extensive experiments on abdominal CT-MRI registration showed that our strategy could promisingly advance the original learning-based method in terms of efficient hyperparameter tuning and a better tradeoff between accuracy and smoothness.

%
%
%
\bibliographystyle{splncs04.bst}
\bibliography{refs.bib}
\end{document}